\documentclass[preprint,12pt]{elsarticle}
\usepackage{amssymb, amsmath, amsfonts}
\usepackage{multirow}           % 표 multi-row
\usepackage[normalem]{ulem}     % 표 underline
\useunder{\uline}{\ul}{}            % 표 underline 명령어
\usepackage{pifont}             % 표 체크, X 표시
\newcommand{\cmark}{\ding{51}}      % 체크
\usepackage{amssymb}            % 수식 intercal
\usepackage{dsfont}             % 수식
\usepackage{graphicx} 
\usepackage[font=small,labelfont=bf]{caption}
\usepackage{subcaption}
\usepackage[hidelinks]{hyperref}
\usepackage{algorithm}
\usepackage{algpseudocode}
\usepackage{array}

\usepackage{xcolor}
\newcommand{\rv}[1]{\textcolor{black}{#1}}

\journal{Knowledge-Based Systems}

\begin{document}

    \newcommand{\mname}{CRKT}
    \newcommand{\mfull}{Concept map-driven Response disentanglement method for enhancing Knowledge Tracing}
    
    \newcommand{\ptitle}{Enhancing Knowledge Tracing with Concept Map and Response Disentanglement}
    \newcommand{\ie}{\textit{i}.\textit{e}.}
    
    \begin{frontmatter}
    
    \title{\ptitle{}}
    
    \author[SKKU]{Soonwook Park}
    \ead{soonwook34@g.skku.edu}
    
    \author[Poly]{Donghoon Lee}
    \ead{leex6962@edu-poly.com}
    
    \author[SKKU]{Hogun Park}
    \ead{hogunpark@skku.edu}
    
    \affiliation[SKKU]{organization={Department of Artificial Intelligence, Sungkyunkwan University},
                addressline={2066, Seobu-ro}, 
                city={Jangan-gu, Suwon-si},
                state={Gyeonggi-do},
                postcode={16419}, 
                country={Republic of Korea}}
    \affiliation[Poly]{organization={Department of AI Development, Poly Inspiration},
                addressline={201, Songpa-daero}, 
                city={Songpa-gu},
                state={Seoul},
                postcode={05854}, 
                country={Republic of Korea}}
    
    \begin{abstract}
    In the rapidly advancing realm of educational technology, it becomes critical to accurately trace and understand student knowledge states. Conventional Knowledge Tracing (KT) models have mainly focused on binary responses (\ie, correct and incorrect answers) to questions. Unfortunately, they largely overlook the essential information in students' actual answer choices, particularly for Multiple Choice Questions (MCQs), which could help reveal each learner's misconceptions or knowledge gaps. To tackle these challenges, we propose the \mfull{} (\mname{}) model. \mname{} benefits KT by directly leveraging answer choices—beyond merely identifying correct or incorrect answers—to distinguish responses with different incorrect choices. We further introduce the novel use of unchosen responses by employing disentangled representations to get insights from options not selected by students. Additionally, \mname{} tracks the student's knowledge state at the concept level and encodes the concept map, representing the relationships between them, to better predict unseen concepts. This approach is expected to provide actionable feedback, improving the learning experience. Our comprehensive experiments across multiple datasets demonstrate \mname{}'s effectiveness, achieving superior performance in prediction accuracy and interpretability over state-of-the-art models. 
    \end{abstract}

    % 표지
    \begin{titlepage}
        \begin{center}
            {\Large \ptitle{}} \\
            \vspace{1cm}
            Soonwook Park,
            Donghoon Lee$^{*}$,
            Hogun Park$^{*}$,
            \\
            \vspace{0.5cm}
            {\footnotesize \textit{Department of Artificial Intelligence, Sungkyunkwan University, 2066, Seobu-ro, Jangan-gu, Suwon-si, Gyeonggi-do, 16419, Republic of Korea}} \\
            {\footnotesize \textit{Department of AI Development, Poly Inspiration, 201, Songpa-daero, Songpa-gu, Seoul, 05854, Republic of Korea}} \\
        \end{center}
        \let\thefootnote\relax\footnote[1]{\footnotesize{$^*$Corresponding Author}}
        \let\thefootnote\relax\footnote[1]{\footnotesize{~~Contact email: leex6962@edu-poly.com (Donghoon Lee) and hogunpark@skku.edu (Hogun Park)}}
    \end{titlepage}
    
    %% Research highlights
    \begin{highlights}
    \item Utilizing option responses for detailed insights into student knowledge states.  % 78자
    \item Introducing unchosen responses to learn the intent of the chosen response in MCQs.  % 82자
    \item Employing concept maps for enhanced interpretability and performance.  % 68자
    \item Demonstrating superior prediction performance over baselines in five datasets.  % 78자
    \item Validating effectiveness in improving personalized learning experiences.  % 71자
    \end{highlights}
    
    
    \end{frontmatter}
    
    %% main text
    %!TEX root = 0_main.tex

\section{Introduction}
% Knolwedge Tracing에 대해
Knowledge Tracing (KT) serves as a critical component in educational technology, capturing the dynamic processes of human learning~\cite{kt_survey, first_kt, bkt}. It outlines a learner's cognitive state through their interaction history with educational materials, predicting future achievements and potential misunderstandings. Recent advances in KT methodologies, particularly through deep learning, have significantly enhanced our ability to understand and predict learning trajectories with remarkable accuracy~\cite{saint, akt, dkt}.

% Knowledge Tracing의 한계: Option Response 활용 X
Despite such progress, conventional KT models still exhibit some limitations. Primarily, those models often rely only on binary responses, \ie{}, correct or incorrect, to infer students' knowledge states. These responses overlook the differences that can be drawn from the multifaceted nature of student responses, especially in the context of Multiple Choice Questions (MCQs)~\cite{ot}. Such questions typically provide various incorrect options, each potentially reflecting different student misconceptions or knowledge gaps. This reliance on binary responses limits the depth and accuracy of insights gained from KT, as it fails to capture the nuances in student responses.

% Knowledge Tracing의 한계: Unchosen Response 활용 X
Furthermore, the valuable data inherent in unchosen responses—options considered but not selected by the student—remains largely unexplored. These unchosen options offer critical insights into students' decision-making processes and conceptual understanding, providing a more comprehensive view of their knowledge states. While a few KT models have leveraged options in the KT scenario~\cite{dp-mtl}, they are still limited in obtaining robust and rich knowledge state representations.

% Knowledge Tracing의 한계: Knolwedge State의 Interpretability 부족
Deep learning models have recently enhanced the capability to track complex patterns in learning data by representing knowledge states in multi-dimensional vectors~\cite{sakt}. While effective in capturing intricate learning trajectories, this approach often obscures the interpretability of knowledge states, making it challenging to provide actionable feedback to learners or educators~\cite{qikt}. In other words, representing knowledge states in multi-dimensional vector spaces does little to clarify specific areas of strength or improvement for individual learners.

% Knowledge Tracing의 한계: Concept Map의 활용 X
Additionally, some existing models acknowledge the relevance of concepts associated with questions but largely neglect the complex relationships between these concepts~\cite{cmkt, dkvmn}. These concepts often serve as prerequisites for others or function within higher categories according to a learning roadmap. The inability of KT models to utilize these proper inter-concept relationships leads to a missed opportunity for achieving a comprehensive and interconnected understanding that extends beyond the concepts directly associated with individual questions.

% 제안 모델 요약
To tackle the aforementioned limitations, we propose the \mfull{} (\mname{}) in this work. This innovative approach leverages detailed student responses to track knowledge states precisely. It also utilizes inter-concept relationships to access students' latent abilities, significantly enhancing interpretability in the process.

% [Figure] 기존 모델들의 한계점
\begin{figure}[!t]
    \centering
    \includegraphics[width=\linewidth]{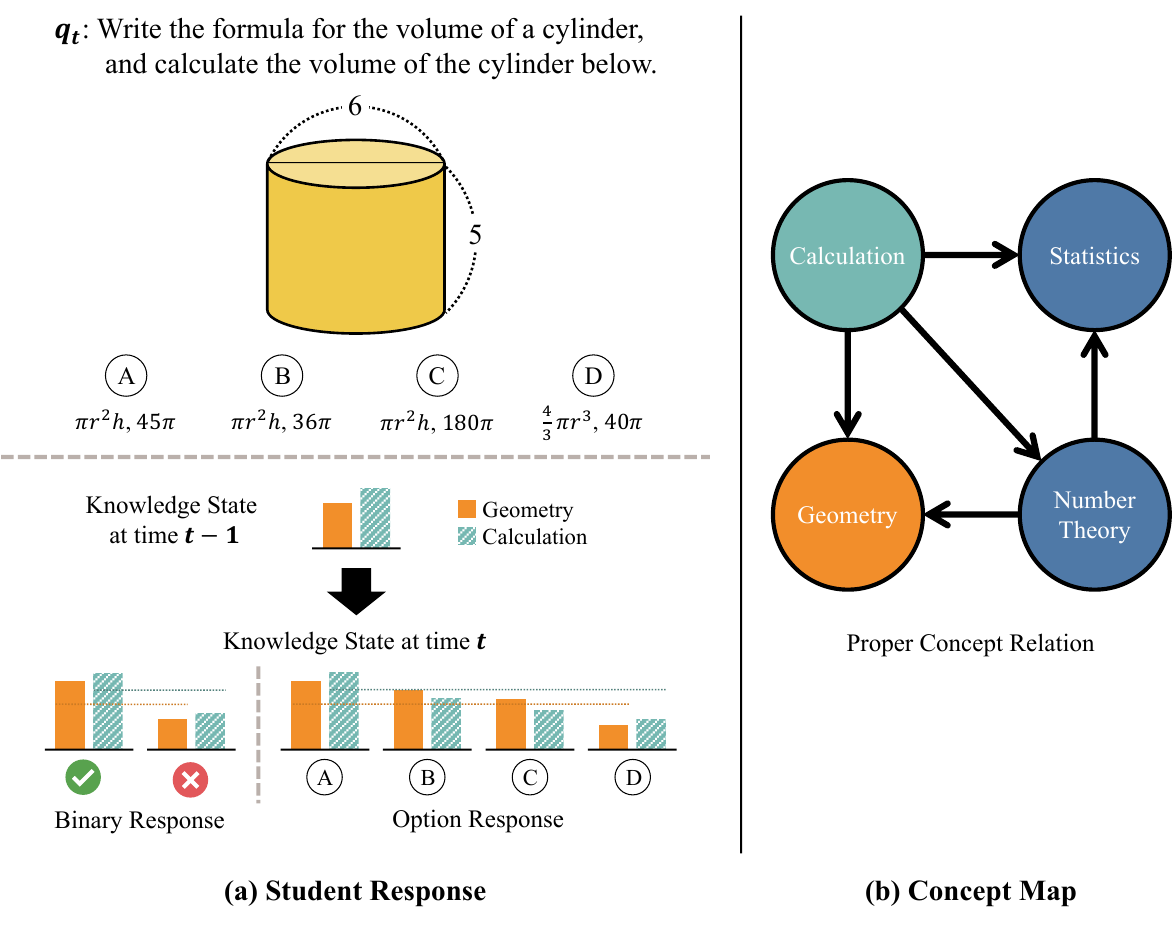}
    \caption{Limitations of conventional KT models: (a) Reliance on binary responses only and (b) Unutilized concept relationships}
    \label{fig:intro}
\end{figure}

% 제안 모델 소개: Option Response 활용
Specifically, \mname{} extends the conventional representation of correct or incorrect responses to questions into an option-based representation to utilize option responses effectively. Binary responses ignore the unique characteristics of each incorrect answer in MCQs. To illustrate this, Fig.~\ref{fig:intro}(a) shows a question where the correct answer is (A), and the incorrect answers are (B), (C), and (D). Conventional KT models have treated all incorrect responses equally; however, each incorrect answer may indicate different concepts the student lacks or specific errors they have made. By leveraging option responses to differentiate among incorrect answers, \mname{} not only improves the accuracy of knowledge state estimation but also enables customized interventions to meet the specific needs of learners.

% 제안 모델 소개: Unchosen Response 활용
In this work, we suggest utilizing both \emph{chosen} and \emph{unchosen} responses to obtain more precise knowledge states. Incorporating such unchosen options allows the KT model to have a more comprehensive understanding of student responses compared to utilizing only chosen option responses~\cite{ot, dp-mtl}. A benefit of the proposed method is manifested in a case study in Fig.~\ref{fig:intro}(a). Here, options (A), (B), and (C) correctly have the formula to compute the volume of a cylinder, while option (D) has the irrelevant formula for the volume of a sphere. If the student chooses option (D), it reflects his/her misunderstanding of geometric concepts and needs to handle them differently from other options that reflect a correct understanding of those. Overall, the proposed method is expected to reveal the reason behind the student response and distinguish each student's strong and weak areas.

% 제안 모델 소개: Knolwedge State의 Interpretability
Furthermore, we enhance the interpretability of knowledge states by transitioning from conventional high-dimensional vector representations to concept-level scalar values. This shift significantly improves the utility of the KT model by making the outputs easier for educators to interpret. These interpretable knowledge states facilitate the provision of detailed, actionable feedback to learners, thereby supporting more effective personalized learning experiences.

% 제안 모델 소개: Concept Map의 활용
Additionally, it is crucial to utilize the relationships among concept-level knowledge states that are obtained by integrating both chosen and unchosen responses. These relationships play a significant role in accurately predicting responses to questions involving unseen concepts. We introduce a graph-based concept map, which effectively captures relationships among concepts. The concept map used in this model is based on predefined relationships determined by experts, ensuring its relevance and effectiveness in enhancing prediction accuracy. Fig.~\ref{fig:intro}(b) provides an example of how the concepts influencing the question in Fig.~\ref{fig:intro}(a) are related. Assuming a student is encountering a geometry question for the first time, using the concept relations allows for accurate predictions by referencing the knowledge state of related concepts like number theory. However, if the relationships among concepts are not utilized, accurate predictions cannot be made for questions involving concepts that a student is encountering for the first time.

To effectively leverage the complex interactions between concepts, we dynamically form question-specific concept maps, considering the required conceptual understanding for each question and the relationships between related concepts. This approach ensures that the knowledge states account for the concepts directly associated with a question and their proper related concepts.

In summary, the main contributions of this work are as follows:
\begin{itemize}
    \item We propose a novel KT model, the \mfull{} (\mname{}), that expands students' binary responses to option responses and applies question-specific concept maps to trace richer knowledge states.
    \item We utilize students' unchosen responses for a more accurate understanding of students' knowledge states. By focusing on why students selected specific options over other alternatives, we identify strengths and weaknesses that are not discernible through chosen responses alone.
    \item We enhance the interpretability of the KT model by employing the concept level knowledge states. This facilitates the actionable feedback design for each learner.
    \item We introduce and utilize dynamic concept maps to capture complex inter-concept relationships effectively, leading to more accurate predictions.
    \item We conducted multiple experiments on four publicly available datasets and one internally developed dataset to validate the superiority of \mname{} over the latest baselines\footnote{Source code is available at \url{https://github.com/Soonwook34/CRKT}}.
\end{itemize}

    %!TEX root = 0_main.tex

\section{Related Work}

% Knowledge Tracing에 대해
\subsection{Knowledge Tracing}
Knowledge Tracing is crucial in educational data mining, offering a methodology to assess and model students' knowledge over time.~\cite{akt, bkt, dkt, dkt-plus} This process predicts the likelihood of a student correctly answering upcoming questions, thus providing a dynamic snapshot of their learning trajectory. KT methodologies have evolved from Item Response Theory models~\cite{irt, irt2}, which utilize user and item parameters to measure question difficulty and student ability, to more sophisticated sequential approaches that capture the subtle differences of learning history~\cite{cl4kt, dtransformer}. Recent advancements have introduced Recurrent Neural Networks (RNN), Dynamic Key-Value Memory Networks (DKVMN), and transformer-based models, significantly broadening the scope and accuracy of KT.

% Knowledge Tracing에서 사용하는 Concept과 그 사이간의 관계
\subsection{Concept Relations in Knowledge Tracing}
The performance enhancements in KT have been achieved through various features, among which concepts have played a crucial role in compressing the representation between numerous questions and sharing knowledge mastery. Recent studies have strived to represent the relationships between these concepts, either directly or indirectly~\cite{cmkt, infosci_graph-cqs, infosci_graph-contrast}. Various approaches have been employed to form these relations, such as viewing the relationships between concepts as a dense graph~\cite{gkt}, creating a joint graph with questions or students through the student's problem-solving history~\cite{rcd, infosci_graph-joint}, and forming a hypergraph with various features like templates~\cite{infosci_graph-hyper}. Additionally, variations of Graph Neural Networks (GNNs) have been widely used to encode these structures efficiently.

However, these studies have limitations in forming the interrelations within the concepts; they directly generate relations using problem-solving history or statistical methods not designed by experts. Therefore, these approaches can potentially create incorrect connections or omit meaningful relationships, acting as noise in tracking the knowledge state.

% Option Tracing에 대해
\subsection{Option Tracing}
Option Tracing (OT) advances KT by analyzing students' choices on MCQs beyond simply whether answers are correct or not~\cite{ot}. Conventional KT methods focus on binary response data, offering a broad estimate of student mastery but neglecting the detailed insights that can be obtained from analyzing incorrect answer choices. Incorrect answers, after all, can result from varied misconceptions or knowledge gaps~\cite{ot-math, ot-wrong}. OT aims to fill this gap by predicting a student's exact option, providing a finer-grained view of their understanding. This approach moves beyond binary correctness to explore the rich information within student answer choices, enhancing assessments by pinpointing specific areas of misunderstanding or lack of knowledge~\cite{p-irt}.

Initial OT models did not integrate with KT in a multi-task learning framework. To address this, recent developments have proposed sophisticated frameworks that handle both KT and OT~\cite{dp-mtl}. This approach enhances the prediction accuracy of KT models by considering the differences in option responses. Nevertheless, it still has significant limitations as it misses valuable insights into knowledge states that came from unchosen responses.
% Nevertheless, initial OT models did not integrate with KT in a multi-task learning setup. Recognizing the limitation, recent developments propose sophisticated frameworks that handle both KT and OT simultaneously, improving the prediction accuracy by considering the differences in option selection~\cite{dp-mtl}.

% Disentangled Representation에 대해
\subsection{Disentangled Representation Learning}
Disentangled learning aims to separately represent underlying factors within complex data~\cite{disenc2, beta-vae, disenc3}. This method organizes information into distinct vector segments, focusing on reducing the overlap or mutual information between these segments, promoting a more precise and interpretable representation of each factor. By doing so, disentangled learning enhances models' accuracy and explainability across various domains~\cite{dis-rec1, disenc1}. The goal is to better understand the data by isolating its components, which can significantly improve decision-making and predictive performance.

In recommendation systems, user preferences are filtered by multi-feedback, including negative and unclicked items, enhancing performance and interpretability~\cite{cdr, dis-rec2}. Inspired by such research, we introduce disentangled representations to extract a more precise knowledge state from the user's responses.
    
    %!TEX root = 0_main.tex

\section{Preliminary}
This section provides an overview of knowledge tracing tasks and introduces essential concepts and terms for this study.

% Knowledge Tracing 기본 정의
\subsection{Problem Statement}
\rv{Knowledge tracing aims to track a student's current knowledge state based on their past problem-solving history and to predict whether a given target question will be answered correctly. For a student $s$, assuming the student $s$ has sequentially solved questions up to time $t$, the problem-solving history $\mathcal{X}$ is denoted as $\mathcal{X} = \{(q_1, y_1), \dots, (q_t, y_t)\}$, where $q_t$ is an index of question at time $t$ and $y_t \in \{0, 1\}$ indicates whether student $s$ answered question $q_t$ correctly. Meanwhile, we can leverage additional information to precisely understand a student's knowledge states. Previous studies have underscored the significance of employing concepts associated with questions. Given that a question may be linked to multiple concepts, the concept set of question $q_t$ is denoted as $\mathcal{C}_{q_t} \subset \mathcal{C}$, where $\mathcal{C}$ represents the entire set of concepts. In addition, the option response $o_t \in \mathcal{O}_{q_t}$ that student $s$ selected can be available in multiple-choice questions, where $\mathcal{O}_{q_t}$ is the set of choices in $q_t$. We also consider options that are not chosen by student $s$. The unchosen response set $\mathcal{U}_t$ can be obtained by removing $o_t$ from the set of choices, $\mathcal{O}_{q_t} - \{o_t\}$. Taking all of this into account, the problem-solving history $\mathcal{X}$ is redefined as follows: $\mathcal{X} = \{(q_1, y_1, \mathcal{C}_{q_1}, o_1, \mathcal{U}_1), \dots, (q_t, y_t, \mathcal{C}_{q_t}, o_t, \mathcal{U}_t)\}$, incorporating not only the questions and their correctness but also the associated concepts, and chosen and unchosen option responses.}

In this study, we aim to address the KT task through a deep learning model $\mathcal{M}$. Consequently, given the problem-solving history $\mathcal{X}$, the KT model $\mathcal{M}$ tracks the knowledge state at time $t$ and predicts the probability $p_{t+1}$ that student $s$ will answer the target question $q_{t+1}$ correctly. This probability can be denoted as $p_{t+1} = p(y_t = 1 | \mathcal{X}, q_{t+1})$.

% Unchosen Response가 어떤식으로 작동하는지
\subsection{Unchosen Response}
In existing research, correct and incorrect responses have been utilized as indicators of student responses to questions~\cite{dkt}. Given KT's goal of deducing a student's knowledge state from a sequence of responses, it logically follows that representations of these responses in KT models reflect the degree of mastery increase or decrease over knowledge elements like concepts. This paper expands upon traditional binary response models by incorporating detailed option responses for each question, characterized by which option was chosen. Thus, the representation of each option response is interpreted as signaling a change in the knowledge state upon the selection of that option.

% [Figure] Unchosen response에 대한 설명
\begin{figure}[!ht]
    \centering
    \includegraphics[width=0.8\linewidth]{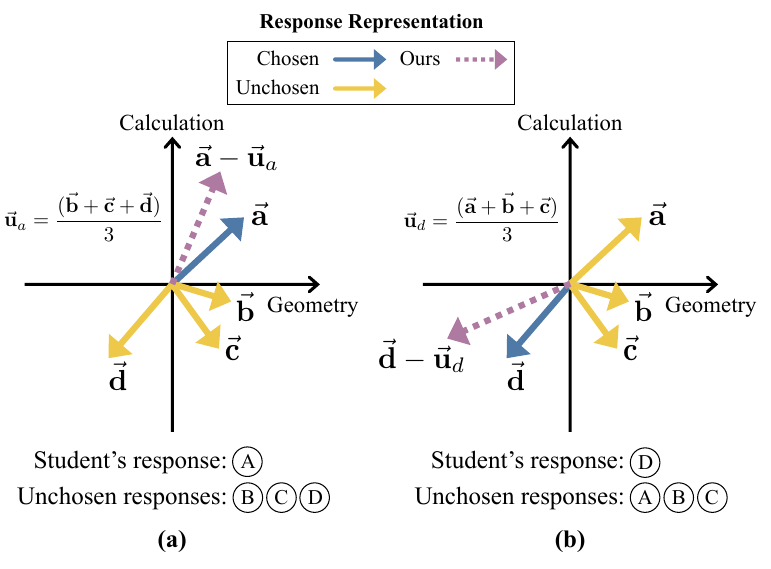}
    \caption{Examples of disentangled response representation for the question in Fig.~\ref{fig:intro}.}
    \label{fig:unchosen}
\end{figure}

Building on the discussion of selection responses, this study also incorporates responses that are not chosen. Under the assumption that a student did not choose an answer due to guessing or mistakenly not following their original intention, a student's decision to exclude specific options in favor of another indicates positive or negative discrimination between chosen and unchosen options. This discrimination offers additional insights into the strengths or weaknesses that the student may perceive in the chosen option relative to the others. We compare the representations of chosen and unchosen options to exploit this discrepancy. For instance, Fig.~\ref{fig:unchosen} depicts the representations for each option response of the question in Fig.~\ref{fig:intro} as a two-dimensional vector encompassing `Geometry' and `Calculation'. Fig.~\ref{fig:unchosen}(a) illustrates a scenario where a student selects the correct option (A). The difference in the option response representations, $\vec{\mathbf{a}} - \vec{\mathbf{u}}_a$, emphasizes the student's calculation proficiency, as option (A) is the only one with the correct calculation. Similarly, Fig.~\ref{fig:unchosen}(b) shows a scenario where a student selects the incorrect option (D). Since all other options (A), (B) and (C) represent the correct formula, the choice of (D) underscores the student's deficiency in understanding geometry. Thus, analyzing unchosen responses enables identifying aspects of students' abilities that are not evident from chosen responses alone.

% Given Concept Map이 없는 경우 만드는 방법에 대해
\subsection{Concept Map} \label{sec_pre_map}
Previous research has explored the connections among concepts by generating graphs from a key-value memory~\cite{cmkt} or employing dense graphs~\cite{gkt}. Such approaches have been deemed valuable for representing knowledge states at the concept level and understanding the interrelations among concepts without treating each concept in isolation. However, directly generating relationships between concepts does not ensure that the connections formed are interpretable and accurate.

Therefore, we have decided to employ concept maps. A concept map is explicitly designed by content managers during the formation of learning data. These maps contain hierarchical curriculum information for the entire learning dataset or only include verified connections. The concept map is denoted as $G = (\mathcal{C}, \mathcal{E})$, where $\mathcal{E}$ is a set of edges $e_{ij}$ where the relation from concept $c_i$ to $c_j$ in the concept set $\mathcal{C}$ exists. 

For relationships not directly specified in the dataset, we introduce a statistical method to find connections between concepts~\cite{rcd}. We first form a correct matrix $\mathbf{V} \in \mathbb{R}^{|\mathcal{C}| \times |\mathcal{C}|}$ from the dataset, where $\mathbf{V}_{i,j} = \frac{n_{ij}}{\sum_k{n_{ik}}}$ is the $(i, j)^{\text{th}}$ entry of $\mathbf{V}$. $n_{ij}$ represents the number of times a question associated with concept $c_i$ is answered right before a question related to concept $c_j$ is correctly answered. Next, we derive a transition matrix $\mathbf{V}^\prime \in \mathbb{R}^{|\mathcal{C}| \times |\mathcal{C}|}$ from $\mathbf{V}$, where $\mathbf{V}^\prime_{i,j} = \frac{\mathbf{V}_{i,j} - \min(\mathbf{V})}{\max(\mathbf{V}) - \min(\mathbf{V})}$ is the $(i, j)^{\text{th}}$ entry of $\mathbf{V}^\prime$. The elements of $\mathbf{V}^\prime$ indirectly denote the probability of a relationship between \(c_i\) and \(c_j\). We used only the connections where $\mathbf{V}^\prime_{i,j} > threshold$, where we set $threshold$ as the average value of $\mathbf{V}^\prime$ in this paper.

% IRT-based Prediction
\subsection{Item Response Theory}
Item Response Theory (IRT)~\cite{irt} is a foundational framework in educational assessment designed to model the relationship between a student's latent ability and their responses to questions. In the mid-20th century, IRT has evolved to provide more accurate insights into learner performance and question attributes than traditional scoring methods. Central to IRT is the Item Characteristic Curve (ICC)~\cite{icc}, a mathematical model that connects a student's ability to the difficulty level of an item. This model posits that the probability of a correct response is influenced by the interaction between the student's ability and specific features of the question, such as its difficulty.

Integrating IRT into deep learning-based KT models enhances interpretability by leveraging the student's history of item responses to predict the likelihood of accurately answering future questions~\cite{rcd}. The ICC, fundamental to IRT, expresses the probability of a student answering a question correctly, based on their ability and the item's difficulty, which can be denoted as:
\begin{equation}
    p(\text{correct response}) = \frac{1}{1 + e^{-(ability - difficulty)}}.
\end{equation}
% Integrating IRT into deep learning-based Knowledge Tracing (KT) models enhances interpretability by leveraging the student's history of item responses to predict the likelihood of accurately answering future questions. The ICC, fundamental to IRT, is represented by a specific equation $p(\text{correct response}) = \frac{1}{1 + e^{-(ability - difficulty)}}$ that expresses the probability of a student answering a question correctly, based on their ability and the item's difficulty.

In our approach, we focus on utilizing the conceptual framework provided by the ICC to improve not only the interpretability of our model's predictions but also to facilitate the provision of feedback in practical applications.

    %!TEX root = 0_main.tex

\section{Method}

% [Figure] Model Overview
\begin{figure*}[!t]
    \centering
    \includegraphics[width=\textwidth]{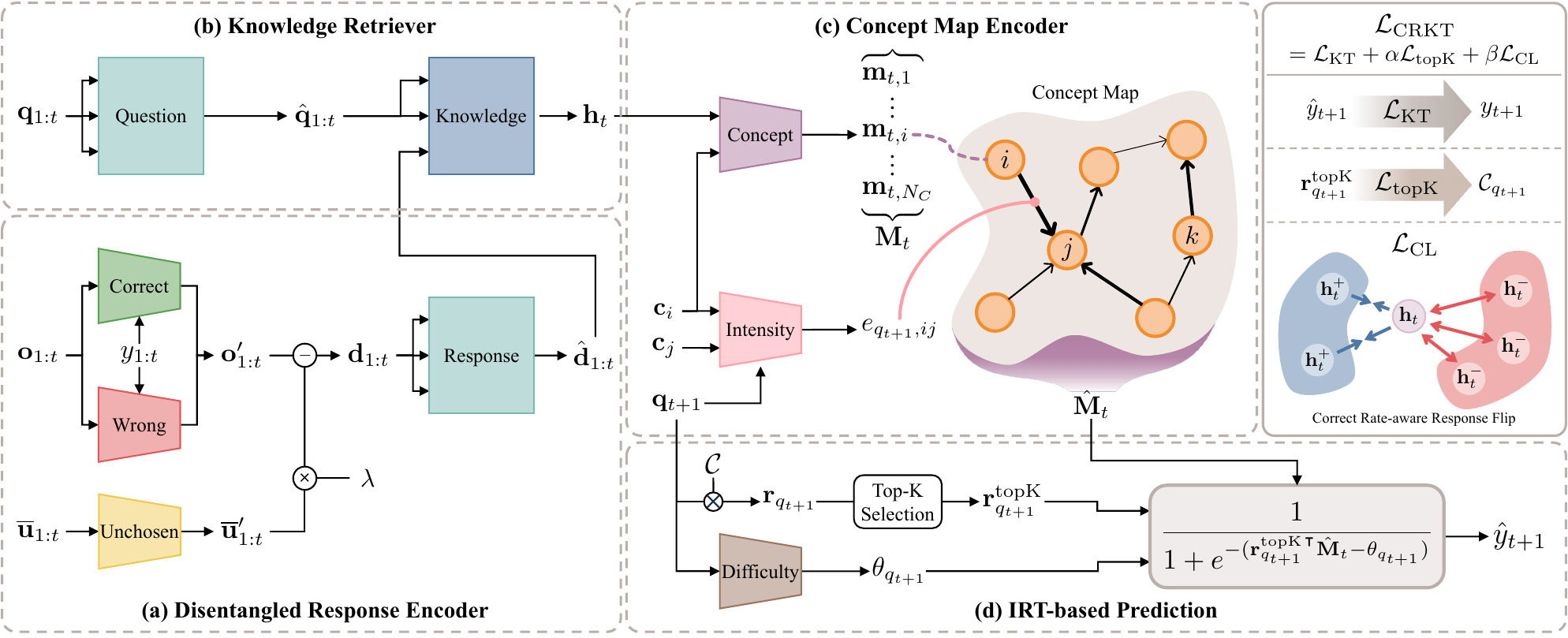}
    \caption{The architecture of \mfull{} model (\mname{}).}
    \label{fig:model}
\end{figure*}

% 모델 간략 소개
This section provides a detailed explanation of the \mname{}. The foundational architecture of \mname{} is illustrated in Fig.~\ref{fig:model}. \mname{} processes the student's selected and unchosen option responses through response encoders, accounting for the context of responses and generating a comprehensive representation of the student's responses. A unified knowledge state is constructed using an attention-based knowledge retriever in conjunction with the question representation. This mechanism is further refined at the concept level, integrating into a concept map with question-specific edge weights. The knowledge state for each concept is then established by applying a GNN~\cite{gcn}, converting each knowledge state into a scalar value. To enhance interpretability in prediction, the relevance of each concept to the target question is calculated, aiding in deducing the student's ability to solve the target question. Finally, predictions about the accuracy of responses are made by incorporating IRT~\cite{irt} to consider the questions' difficulty.

% 모듈 소개: Disentangled Response Encoder
\subsection{Disentangled Response Encoder}
A student's knowledge state at time $t$ can be determined through his history of item responses from time $1$ to $t$. We aim to compile a student's ultimate response representation by incorporating his chosen and unchosen responses. First, the encoding of a student's chosen response is as follows:
\begin{equation}
    \label{eq:enc_chosen}
    \mathbf{o}^\prime_t = 
    \begin{cases}
        f_{\text{correct}}(\mathbf{o}_t) & \text{if $y_t = 1$}\\
        f_{\text{wrong}}(\mathbf{o}_t) & \text{if $y_t$ = 0},
    \end{cases}
\end{equation}
where $\mathbf{o}_t \in \mathbb{R}^{d_q}$ represents the $d_q$-dimensional embedding vector for the student's selected response $o_t$, which undergoes distinct encoding processes based on its correctness $y_t$. $f_{\text{correct}}$ and $f_{\text{wrong}}$ are both 2-layer MLPs (Multi-Layer Perceptrons) that generate a correctness-aware representation $\mathbf{o}^\prime_t \in \mathbb{R}^{d_q}$.

Next, we address the encoding of a student's unchosen responses. While the chosen response provides direct insight into the student's knowledge, the unchosen responses can reveal additional nuances about their understanding and misconceptions. To incorporate the influence of these unchosen responses, we derive their representation as follows:
\begin{equation}
    \label{eq:enc_unchosen}
    \overline{\mathbf{u}}^\prime_t = f_{\text{unchosen}}(\overline{\mathbf{u}}_t),
\end{equation}
where $\overline{\mathbf{u}}_t$ denotes the average embedding vector for the unchosen options, which is the mean of all option response representations in the unchosen set $\mathcal{U}_t$. This embedding vector shares the embedding space with the chosen response $\mathbf{o}_t$. To signify its status as unchosen by the student, $\overline{\mathbf{u}}_t$ is processed through a 2-layer MLP encoder $f_{\text{unchosen}}$.

To integrate the insights gained from both the chosen and unchosen responses, we derive a disentangled representation of the student's response as follows:
\begin{equation}
    \label{eq:enc_disentangle}
    \mathbf{d}_t = \mathbf{o}^\prime_t - \lambda \cdot \overline{\mathbf{u}}^\prime_t.
\end{equation}
In this equation, $\mathbf{d}_t$ represents the disentangled response representation. It is calculated by subtracting the encoded unchosen response $\lambda \cdot \overline{\mathbf{u}}^\prime_t$ from the encoded chosen response $\mathbf{o}^\prime_t$. $0<\lambda < 1$ is a scaling factor that adjusts the influence of the unchosen response. This approach allows us to isolate the student's decision-making process by comparing the chosen response with the average representation of the unchosen responses, thereby highlighting the specific knowledge or misconceptions that influenced their choice.

Upon obtaining the sequence representations, we utilize a self-attention mechanism to integrate information spanning from time $1$ to $t$. We adopt the temporal and cumulative attention mechanism, a proven technique that ensures context-aware representations~\cite{dtransformer}. This approach is thoroughly detailed in Section~\ref{sec_kr}. Consequently, the final representation of a student's item responses up to time $t$, $\hat{\mathbf{d}_t}$, can be obtained.

% 모듈 소개: Knowledge Retriever
\subsection{Knowledge Retriever} \label{sec_kr}
Attention-based methods offer distinct advantages over traditional sequence-based KT approaches. One of the primary strengths of attention-based methods is their ability to assign different levels of importance to past response records by applying weights that consider the target question. We aim to retrieve knowledge states by integrating the question representation sequence with the disentangled response representation sequence. Inspired by DTransformer~\cite{dtransformer}, we utilize a temporal and cumulative attention mechanism that leverages the following attention calculation: 
\begin{equation}
    \label{eq:enc_attn}
    \text{AttentionMax}(\mathbf{Q}, \mathbf{K}, \mathbf{V}) = \text{MaxOut}(\frac{\mathbf{Q}\mathbf{K}^\intercal \cdot e^{-\eta \cdot d(\Delta{t})}}{\sqrt{d_{k}}})\mathbf{V}
\end{equation}
with
\begin{equation}
    \label{eq:enc_maxout}
    \text{MaxOut}(x_i) = \frac{\text{softmax}(x_i)}{\max_j\{\text{softmax}(x_j)\}},
\end{equation}
where $\eta$ is the parameter that controls the strength of the temporal effect, and $d(\Delta{t})$ is the distance function measuring the temporal and semantic distance between a query and a key. This approach helps aggregate relevant experiences while considering the learning process's temporal aspects. The MaxOut operation ensures that repeated efforts on the same concept or question are properly weighted, reflecting the cumulative effect of learning. By combining temporal and cumulative considerations, this attention mechanism allows the DTransformer to estimate the knowledge state better.

The specific implementation details of this module are outlined in Algorithm~\ref{alg:attn}. Similar to the disentangled response representation $\mathbf{d}_t$ on line 10, the question representation $\mathbf{q}_t \in \mathbb{R}^{d_q}$ is also subjected to a self-attention mechanism to derive context-aware representation $\hat{\mathbf{q}}_t$ on line 9 in Algorithm~\ref{alg:attn}.

The knowledge retriever also employs the temporal and cumulative attention mechanism to discover a student's knowledge state $\mathbf{h}_t$ at time $t$, using the sequence of questions as both the query and key, and the sequence of responses as the value on line 12 in Algorithm~\ref{alg:attn}. This approach ensures that the derived knowledge state is closely aligned with the contextual relevance of the question and the student's cumulative response pattern, enabling a precise evaluation of the student's current understanding.

\begin{algorithm}[t]
    \color{black}
    \caption{Knowledge Retriever}\label{alg:attn}
    \begin{algorithmic}[1]
        \Function{AttentionMax}{$\mathbf{Q}$, $\mathbf{K}$, $\mathbf{V}$} // $\mathtt{(Eq. 5)}$
            \State \textbf{Input} query $\mathbf{Q}$, key $\mathbf{K}$, value $\mathbf{V}$
            \State \textbf{return} $\text{MaxOut}(\dfrac{\mathbf{Q}\mathbf{K}^\intercal \cdot e^{-\eta \cdot d(\Delta{t})}}{\sqrt{d_{k}}})\mathbf{V}$
        \EndFunction
        \State
        \Function{KnowledgeRetriever}{$\mathbf{q}_{1:t}$, $\mathbf{d}_{1:t}$}
            \State \textbf{Input} sequence of question embedding $\mathbf{q}_{1:t}$, sequence of disentangled response embedding $\mathbf{d}_{1:t}$
            \For{$i = 1$ to $t$}
                \State $\hat{\mathbf{q}}_i = \Call{AttentionMax}{\mathbf{q}_{1:i}, \mathbf{q}_{1:i}, \mathbf{q}_{1:i}}$
                \State $\hat{\mathbf{d}}_i = \Call{AttentionMax}{\mathbf{d}_{1:i}, \mathbf{d}_{1:i}, \mathbf{d}_{1:i}}$
            \EndFor
            \State $\mathbf{h}_t = \Call{AttentionMax}{\hat{\mathbf{q}}_{1:t}, \hat{\mathbf{q}}_{1:t}, \hat{\mathbf{d}}_{1:i}}$
        \EndFunction
    \end{algorithmic}
\end{algorithm}

% 모듈 소개: Concept Map Encoder
\subsection{Concept Map Encoder}
To leverage the relationships between concepts and improve interpretability, we first extend the knowledge state to the concept level for mapping onto a concept map as follows:
\begin{equation}
    \label{eq:map_concept-all}
    \mathbf{M}_t = \{\mathbf{m}_{t, 1}, \dots, \mathbf{m}_{t, |\mathcal{C}|}\},
\end{equation}
where
\begin{equation}
    \label{eq:map_concept-single}
    \mathbf{m}_{t, i} = f_{\text{concept}}([\mathbf{h}_t \oplus \mathbf{c}_i]).
\end{equation}
Here $|\mathcal{C}|$ is the number of concepts in $\mathcal{C}$, $\mathbf{c}_i \in \mathbb{R}^{d_c}$ is the $d_c$-dimensional embedding vector of concept $c_i$, and $\oplus$ is the concatenation operation. To formulate the knowledge state related to concept $c_i$, denoted as $\mathbf{m}_{t, i} \in \mathbb{R}^{d_g}$, we concatenate corresponding concept representation $\mathbf{c}_i$ with the aggregated knowledge state $\mathbf{h}_t$ extracted from the student's item response record. This concatenation is processed through a 2-layer MLP, $f_{\text{concept}}$, which transforms a $(d_q + d_c)$-dimensional input into a $d_g$-dimensional representation, thereby yielding $\mathbf{m}_{t, i}$. Accordingly, the collective knowledge state for all $|\mathcal{C}|$ concepts at time $t$, represented as $\mathbf{M}_t \in \mathbb{R}^{|\mathcal{C}| \times d_g}$, is obtained, enabling a comprehensive mapping of concept-related knowledge states to the concept map.

In the given concept map, connections are not treated uniformly; rather, we assign edge weights influenced by the target question $q_{t+1}$ as follows:
\begin{equation}
    \label{eq:map_edge}
    e_{q_{t+1},ij} = \text{ReLU}(f_\text{intensity}([\mathbf{q}_{t+1} \oplus \mathbf{c}_i \oplus \mathbf{c}_j])).
\end{equation}
This adjustment is performed exclusively for edges $e_{ij}$ already present in the concept map's edge set $\mathcal{E}$. The process involves concatenating the representation of the target question $\mathbf{q}_{t+1}$, the source concept $\mathbf{c}_i$, and the target concept $\mathbf{c}_j$ to construct a target question-specific edge weight $e_{q_{t+1},ij} \in \mathbb{R}^1$. A 2-layer MLP $f_\text{intensity}$ is designed to produce a scalar output from an input of dimension $(d_q + 2 \times d_c)$, with $\text{ReLU}$ serving as the activation function. 

After performing this process for all relevant edges, we transform them into an adjacency matrix for practical use. When $\mathbf{A}$ is acknowledged as the adjacency matrix corresponding to the original concept map, the target question-specific adjacency matrix is defined as $\mathbf{A}_{q_{t+1}} \in \mathbb{R}^{|\mathcal{C}| \times |\mathcal{C}|}$. This methodology ensures the dynamic reflection of each edge's relevance to the target question $q_{t+1}$ and facilitates the creation of a question-specific structural representation for enhanced analytical utility. $\mathbf{A}_{q_{t+1}}$ can be normalized as follows:
\begin{equation}
    \label{eq:map_A}
    \tilde{\mathbf{A}}_{q_{t+1}} = \hat{\mathbf{D}}^{-\frac{1}{2}}_{q_{t+1}}\hat{\mathbf{A}}_{q_{t+1}}\hat{\mathbf{D}}^{-\frac{1}{2}}_{q_{t+1}},
\end{equation}
where $\hat{\mathbf{A}}_{q_{t+1}} = \mathbf{A}_{q_{t+1}} + \mathbf{I}$, $\hat{\mathbf{D}}_{q_{t+1}}$ is the degree matrix of $\hat{\mathbf{A}}_{q_{t+1}}$, and $\tilde{\mathbf{A}}_{q_{t+1}}$ is the normalized adjacency matrix. 

To harness the concept map's information, we employ a GNN~\cite{gcn} to update the knowledge state $\tilde{\mathbf{A}}$. The propagation rule for our GNN layer is given by:
\begin{equation}
    \label{eq:map_gcn}
    \mathbf{M}^{(l+1)}_t = \sigma(\tilde{\mathbf{A}}_{q_{t+1}}\mathbf{M}^{(l)}_t\mathbf{W}^{(l)}_m),
\end{equation}
where $\mathbf{W}^{(l)}_m$ denotes the weight matrix that updates the output $\mathbf{M}^{(l+1)}_t$ from the output of $l^{\text{th}}$ layer $\mathbf{M}^{(l)}_t$. For the first GNN layer, we initialize $\mathbf{M}^{(0)}_t = \mathbf{M}_t$.

The GNN architecture encompasses $L$ layers, with the final layer output $\hat{\mathbf{M}}_t = \mathbf{M}^{(L)}_t \in \mathbb{R}^{|\mathcal{C}| \times 1}$ serving as the updated knowledge state. By configuring the output dimension of the last layer to $1$, the knowledge state of each concept is represented as a scalar value, effectively quantifying the student's mastery level in each respective concept.

% 모듈 소개: IRT-based Prediction
\subsection{IRT-based Prediction}
We do not merely feed the knowledge state and target question into a final MLP to get the probability of a student accurately answering the target question. Instead, we incorporate the IRT methodology, significantly enhancing the interpretability of our model's output. This method, inspired by the principles of Multidimensional IRT (M-IRT)~\cite{mirt}, is detailed as follows:
\begin{equation}
    \label{eq:map_pred}
    \hat{y}_{t+1} = \frac{1}{1 + e^{-({\mathbf{r}_{q_{t+1}}^{\text{topK}}}^\intercal\hat{\mathbf{M}}_t - \theta_{q_{t+1}})}},
\end{equation}
where $\mathbf{r}_{q_{t+1}}^{\text{topK}}$ signifies the normalized relevance of the top-$k$ concepts to the target question, and $\theta_{q_{t+1}}$ represents the difficulty level of the question. Limiting the computation to the top-$k$ relevant concepts reduces noise from less relevant concepts, leading to a more accurate estimation of the student's ability. The computation of $\mathbf{r}_{q_{t+1}}^{\text{topK}}$ is described by the subsequent equations:
\begin{equation}
    \label{eq:map_rel-all_topk}
    \mathbf{r}_{q_{t+1}}^{\text{topK}} = \text{softmax}(\{r_{q_{t+1}, c_1}^{\text{topK}}, \dots, r_{q_{t+1}, c_{|\mathcal{C}|}}^{\text{topK}}\}),
\end{equation}
where
\begin{equation}
    \label{eq:enc_rel_topk}
    r_{q_{t+1}, c_i}^{\text{topK}} = 
    \begin{cases}
        r_{q_{t+1}, c_i} & \text{if $r_{q_{t+1}, c_i} \geq b$},\\
        -\infty & \text{otherwise},
    \end{cases}
\end{equation}
and $b$ is the $k^{\text{th}}$ largest value in $\mathbf{r}_{q_{t+1}}$. The relevance score $r_{q_{t+1}, c_i}$ is calculated as follows:
\begin{equation}
    \label{eq:map_rel-single}
    r_{q_{t+1}, c_i} = \sigma({\mathbf{q}_{t+1}}^\intercal(\mathbf{W}_r\mathbf{c}_i)),
\end{equation}
where $\mathbf{q}_{t+1}$ is the representation of the target question and $\mathbf{c}_i$ is the representation of each concept. The relevance is initially derived from the dot product between the representation of the target question $\mathbf{q}_{t+1}$ and the concept representation $\mathbf{c}_i$, adjusted in vector dimension through a weight matrix $\mathbf{W}_r \in \mathbb{R}^{d_q \times d_c}$. After selecting the top-$k$ relevant concepts, this relevance is normalized using the softmax function. By conducting a dot product of $\mathbf{r}_{q_{t+1}}^{\text{topK}}$ with $\hat{\mathbf{M}}_t$, we can obtain a singular question-specific ability from the $|\mathcal{C}|$-dimensional knowledge state. The difficulty of the question is ascertained through the equation:
\begin{equation}
    \label{eq:map_diff}
    \theta_{q_{t+1}} = \sigma(f_{\text{difficulty}}(\mathbf{q}_{t+1})). 
\end{equation}
The difficulty parameter of the target question $\theta_{q_{t+1}} \in \mathbb{R}^{1}$ is derived using a 2-layer MLP $f_{\text{difficulty}}$ that processes the target question $\mathbf{q}_{t+1}$ as its input. This approach allows us to directly correlate the student's conceptual understanding with the inherent challenge of the question, offering a more precise and interpretable prediction of student performance.

\subsection{Model Training}
Finally, to optimize the predicted probability $\hat{y}$ of students correctly answering the target question at all times $T$, we apply the binary cross-entropy loss:
\begin{equation}
    \label{eq:loss_kt}
    \mathcal{L}_{\text{KT}} = -\displaystyle\sum_{t=1}^T{(y_t\log{\hat{y}_t} + (1 - y_t)\log{(1 - \hat{y}_t}))}.
\end{equation}

Additionally, we introduce two more loss functions to enhance the model's performance. The first additional loss function is designed to encourage the relevance score $r_{q_{t+1}, c_i}$ to correctly identify the top-$k$ concepts related to the target question $q_{t+1}$:
\begin{equation}
    \label{eq:loss_topk}
    \begin{split}
        \mathcal{L}_{\text{topK}} = -\displaystyle\sum_{t=1}^T\displaystyle\sum_{i=1}^{|\mathcal{C}|} & \Big( g_{q_{t+1}, c_i}\log{r_{q_{t+1}, c_i}} \\
        & + (1 - g_{q_{t+1}, c_i})\log{(1 - r_{q_{t+1}, c_i})} \Big)
    \end{split}
\end{equation}
with
\begin{equation}
    \label{eq:map_tag-single}
    g_{q_{t+1}, c_i} = \mathds{1}\bigl[c_i \in \mathcal{C}_{q_{t+1}}\bigr].
\end{equation}
This loss function ensures that the concepts most relevant to the target question are ranked higher, thereby improving the relevance and interpretability of the model's predictions. We applied weight to the positive label when conducting experiments to address the imbalance between positive and negative labels. The weight of the positive label is $(|\mathcal{C}| - |\mathcal{C}_{q_{t+1}}|) / k$, where $\mathcal{C}$ represents the entire set of concepts and $\mathcal{C}_{q_{t+1}}$ is the concept set of question $q_{t+1}$.

% The second additional loss function employs contrastive learning to refine the student's knowledge state $\mathbf{h}_t$ using response flip augmentation. This technique generates positive and negative knowledge states by flipping the responses in the student's history based on the average correct rate of the questions:
% The second additional loss function is for contrastive learning applied to the student’s knowledge state $h_t$. We focus on questions with an average correct rate between 40\% and 60\%, where approximately half of the students answer correctly or incorrectly. These questions are notably challenging for KT models to predict, likely because KT models are more influenced by the average correct rate of the questions rather than the students’ actual knowledge states. By accurately distinguishing students’ knowledge states when solving these questions, our model achieves a more refined representation of knowledge states. We employ response flip augmentation on students’ history to generate positive and negative knowledge states to accomplish this.

The second additional loss function is for contrastive learning applied to the student’s knowledge state, with a specific focus on questions that have an average correct rate of 40\% to 60\%. These questions are notably challenging for KT models to predict, largely because KT models' predictions tend to overfit to the average correct rate of the questions. To address this issue, we aim to provide a stronger supervision signal to the knowledge states when students solve these questions, enhancing the differentiation between the knowledge states of students who answer correctly and those who do not. We propose a correct rate-aware response flip augmentation to generate positive and negative histories for each student. From these modified histories, we derive corresponding positive $\mathbf{h}^{s+}$ and negative $\mathbf{h}^{s-}$ knowledge states. Through contrastive learning, we train our model to align the student's original knowledge state $\mathbf{h}^s$ close to $\mathbf{h}^{s+}$ and further from $\mathbf{h}^{s-}$.

Response flip augmentation is an established technique previously implemented through random flips of responses~\cite{dtransformer}. In this paper, we refine the approach by using the average correct rate of questions as a guiding metric. To generate positive/negative samples, we first select questions with a correct rate ranging from 40\% to 60\% as our base questions. 
For students who correctly answer the base question, we flip responses to `correct' for base-related questions with a lower correct rate than the base, resulting in a positive knowledge state that embodies a higher base question-specific ability. Conversely, we flip responses to `wrong' for base-related questions with a higher correct rate than the base, creating a negative knowledge state that reflects a lower ability to the base question. In case of a wrong response to the base question, the criteria for assigning positive and negative samples are inverted.

The contrastive learning loss we employ is defined as follows:
\begin{equation}
    \label{eq:loss_cl}
    \mathcal{L}_{\text{CL}} = -\log{\frac{\exp(\mathtt{sim}(\mathbf{h}^{s}, \mathbf{h}^{s+}))}{\exp(\mathtt{sim}(\mathbf{h}^{s}, \mathbf{h}^{s+})) + \sum_{\mathbf{h}^{s-} \in \mathcal{N}^{s}}{\exp(\mathtt{sim}(\mathbf{h}^{s}, \mathbf{h}^{s-}))}}},
\end{equation}
where $\mathtt{sim}$ denotes cosine similarity, and $\mathbf{h}^{s+}$ and $\mathbf{h}^{s-}$ represent the positive and negative knowledge states obtained through response flip augmentation, respectively. $\mathcal{N}^{s}$ is the set of negative knowledge states.
Furthermore, if there are multiple base questions within the same concepts, the response flip is based on the most recently answered base question. Additionally, a response flip rate is introduced to determine which responses will be flipped randomly. In our experiment, we set the response flip rate to 0.8.

The combined loss function for \mname{} is thus:
\begin{equation}
    \label{eq:loss}
    \mathcal{L}_{\text{CRKT}} = \mathcal{L}_{\text{KT}} + \alpha\mathcal{L}_{\text{topK}} + \beta\mathcal{L}_{\text{CL}},
\end{equation}
where $\alpha$ and $\beta$ are coefficients. This comprehensive loss function trains \mname{}, ensuring a balanced optimization that incorporates predictions' accuracy, concepts' relevance, and knowledge state representations' robustness.

% 모듈 소개: Time Complexity Analysis
\subsection{Time Complexity Analysis}
The \mname{} model is composed of four distinct modules: the disentangled response encoder, knowledge retriever, concept map encoder, and IRT-based prediction module. The disentangled response encoder initially processes the accuracy of selected responses over time, necessitating a time complexity of $O(T)$, where $T$ is the total length of response history. Given that the sequence of responses serves as query, key, and value in the attention mechanism, the resultant time complexity for this component escalates to $O(T^2)$. In the knowledge retriever module, attention mechanisms are similarly employed to a sequence of questions, which results in a complexity of $O(T^2)$. This module also calculates the knowledge state with a time complexity of $O(T)$ because the query is the question at time $t$ while the key and value are the question and response sequence, respectively. The concept map encoder identifies the knowledge state for all concepts. It computes the weight of each edge, leading to a complexity of $O(|\mathcal{C}| + |\mathcal{E}|)$, where $|\mathcal{C}|$ is a number of concepts in $\mathcal{C}$ and $|\mathcal{E}|$ is a number of edges in $\mathcal{E}$. Encoding the concept map through GNN has a complexity of $O(|\mathcal{C}|^2)$, resulting in a total time complexity for the concept map encoder of $O(|\mathcal{C}| + |\mathcal{E}| + |\mathcal{C}|^2)$. The IRT-based prediction module has a time complexity of $O(|\mathcal{C}|)$, calculating the relevance between the targeted question and each concept. 

To summarize, assuming that $O(|\mathcal{E}|) = O(|\mathcal{C}|^2)$, the overall time complexity of \mname{} is articulated as $O(2(T + T^2) + 2|\mathcal{C}| + |\mathcal{C}|^2 + |\mathcal{E}|) \approx O(T^2 + |\mathcal{C}|^2)$. The values of $T$ and $|\mathcal{C}|$ are usually not excessively large. Among the five datasets we experimented with, the largest $|\mathcal{C}|$ was 189, and $T$ was limited to a maximum of 200 through preprocessing.

During the inference stage, the model does not need to continually recompute the attention mechanism for accumulating records. By storing previously computed attention weights for responses and questions up to the $t^{\text{th}}$ question, we can reduce the complexity from $O(T^2)$ to $O(T)$. Similarly, storing pre-computed target-specific edge weights eliminates the need for repeated calculations. This optimization reduces the overall time complexity to $O(T + |C|^2)$, significantly enhancing the model’s efficiency in real-time applications.

    %!TEX root = 0_main.tex

\section{Experiment} % TODO: Numerical tests?
In this section, we conduct experiments to answer the following five research questions: \\
\textbf{RQ1} How does \mname{} perform compared to existing KT models? \\
\textbf{RQ2} How do different components affect the performance of \mname{}? \\
\textbf{RQ3} How do text-based representations compare to general learning-based models? \\
% \textbf{RQ4} In what aspects does \mname{} show performance improvements over existing KT models? \\
\textbf{RQ4} Where does \mname{} improve over existing KT models? \\
\textbf{RQ5} How can \mname{} be applied in real-world scenarios?

% 데이터셋 설명
\subsection{Datasets}
In this section, we describe five datasets used for our experiments. 
Our method applies only to datasets that include students' option responses and concept maps. 
Table~\ref{tab:dataset} presents the statistics for each dataset.

\begin{table}[!h]
    \centering
    \caption{Statistics of Datasets}
    \label{tab:dataset}
    \renewcommand{\arraystretch}{1.2}
    \scalebox{0.9}{
        % \begin{tabular}{c|ccccc}
        \begin{tabular}{c|cc >{\color{black}\arraybackslash}c cc}
            \hline
            Dataset & Poly & DBE-KT22 & EdNet & NIPS34 & ENEM \\
            \hline
            Students & 540 & 1,264 & 5,000 & 4,918 & 10,000 \\
            Concepts & 103 & 93 & 189 & 86 & 185 \\
            Questions & 598 & 212 & 12,044 & 948 & 185 \\
            Options per Question & 4 & 2 $\sim$ 5 & 4 & 4 & 5 \\
            Interactions & 96,615 & 161,952 & 590,925 & 1,382,727 & 1,793,106 \\
            Relations & 101 & 172 & 225 & 85 & 211 \\
            \hline
            Avg. Concepts & 5 & 1.9 & 2.27 & 4.02 & 1 \\
            Correct Rate & 83.39\% & 76.45\% & 66.04\% & 53.73\% & 33.73\% \\
            Sparsity & 70.08\% & 39.56\% & 99.02\% & 70.34\% & 3.08\% \\
            \hline
        \end{tabular}
    }
\end{table}

\begin{itemize}
    \item \textbf{Poly}\footnote{\url{https://www.koreapolyschool.com/main.do}}: The Poly dataset includes questions and students' responses collected from monthly tests in elementary school English grammar courses at Poly learning centers from April 2019 to December 2021. Each test has thirty questions and is aligned with curriculum progression. Furthermore, this dataset features the tree-structured concept map, where each question is tagged from the root node to the leaf node.
    \item \textbf{DBE-KT22}\footnote{\url{https://dataverse.ada.edu.au/dataset.xhtml?persistentId=doi:10.26193/6DZWOH}}~\cite{dbe_kt22}: The Database Exercises for Knowledge Tracing (DBE-KT22) originates from the Database Systems course at the Australian National University. It presents a substantial data source to support and advance research within the KT field. It includes diverse question attributes and provides a concept map with several subgraphs.
    \item \textbf{EdNet}\footnote{\url{https://github.com/riiid/ednet}}~\cite{ednet}: EdNet dataset is collected from Santa, an artificial intelligence tutoring system that aids students in preparing for the TOEIC (Test of English for International Communication). For our experiments, we randomly selected $5,000$ students from the dataset. As no predefined concept map was available, we constructed a concept map using the statistical method outlined in Section~\ref{sec_pre_map}.
    \item \textbf{NIPS34}\footnote{\url{https://eedi.com/projects/neurips-education-challenge}}~\cite{nips34}: This dataset is from the Tasks 3 \& 4 at the NeurIPS 2020 Education Challenge. It contains crowdsourced diagnostic mathematics questions for primary to high school students collected from the educational platform Eedi. It features a concept map with a tree structure like the Poly dataset.
    \item \textbf{ENEM}\footnote{\url{https://github.com/godtn0/DP-MTL/tree/main/data/enem_data}}~\cite{enem}: This dataset is made up of 185 questions from the 2019 Exame Nacional do Ensino Médio (ENEM) in Brazil. Given the lack of predefined concepts and a concept map for this dataset, we treated the questions themselves as concepts. We constructed a concept map employing the statistical method outlined in Section~\ref{sec_pre_map}.
\end{itemize}

% Baseline
\subsection{Baselines}
For an exhaustive comparison with our proposed model, \mname{}, we selected seven KT models as baselines. The descriptions of the baseline models are provided below.

\begin{itemize}
    \item \textbf{DKT}~\cite{dkt}: employs an LSTM layer to encode the students' knowledge state, enabling the prediction of students' response performances.
    \item \textbf{GKT}~\cite{gkt}: is pioneering in integrating a graph into KT by initializing a concept map optimized through predicting responses. It maintains a knowledge state vector for every concept, enabling it to compute the probability of correctly responding to questions linked to a particular concept via the corresponding knowledge state.
    \item \textbf{SAKT}~\cite{sakt}: adopts a simple attention mechanism to assign weights to the significance of previously answered questions within a student's history. 
    \item \textbf{AKT}~\cite{akt}: integrates the monotonic attention mechanism with Rasch model-based embeddings. It incorporates exponentially decaying attention weights based on the contextual distance between questions to address the forgetting effect experienced by students.
    \item \textbf{CL4KT}~\cite{cl4kt}: introduces a general contrastive learning framework for KT. It identifies semantically similar or dissimilar problem-solving histories and captures more effective representations of knowledge states.
    \item \textbf{DTransformer}~\cite{dtransformer}: is a recent method that demonstrates state-of-the-art performance in knowledge tracing. The model features an architecture spanning from question level to knowledge level, explicitly diagnosing the proficiency of students' knowledge from the mastery of each question.
    \item \textbf{DP-DKT}~\cite{dp-mtl}: proposes an integrated framework that combines the traditional KT loss with an option tracing loss to predict the selected option responses in multiple-choice questions accurately. Due to differences in dataset configurations, instead of using Bi-LSTM in DP-BiDKT, DKT incorporating options~\cite{ot} is employed. Accordingly, we changed the name of the model from DP-BiDKT to DP-DKT.
\end{itemize}

% Comparison Table
\begin{table}[!h]
    \centering
    \caption{Feature Usage Comparison}
    \label{tab:usage}
    \renewcommand{\arraystretch}{1.2}
    \scalebox{0.9}{
        \begin{tabular}{l|ccccc}
            \hline
            \multicolumn{1}{c|}{\multirow{2}{*}{Model}} & \multicolumn{5}{c}{Feature} \\
            \cline{2-6} 
            \multicolumn{1}{c|}{} & Concept & Question & Graph & Option & Unchosen \\
            \hline
            GKT &\cmark &- &\cmark &- &- \\
            DKT &\cmark &- &- &- &- \\
            SAKT &\cmark &- &- &- &- \\
            AKT &\cmark &\cmark &- &- &- \\
            CL4KT &\cmark &\cmark &- &- &- \\
            DTransformer &\cmark &\cmark &- &- &- \\
            DP-DKT &\cmark &\cmark &- &\cmark &- \\
            \hline
            \mname{} (ours) &\cmark &\cmark &\cmark &\cmark &\cmark \\
            \hline
        \end{tabular}
    }
\end{table}

To clarify how our proposed model differs from the baseline models, we distinguish them in terms of input features in Table~\ref{tab:usage}. 
% In Table~\ref{tab:usage}, we distinguish our proposed model from the baseline models in terms of input features to clarify how it differs from them. 

\subsection{Experimental Setup}
We set the maximum length of the input sequence to 200 and conducted 5-fold cross-validation for all models and datasets. Additionally, we exclude sequences shorter than $5$ that are too short to extract meaningful knowledge states. We use $80\%$ of the student sequences for training and validation, with the remaining $20\%$ for testing. All models and experiments were implemented using the PyTorch library with version 2.3. We employ the ADAM optimizer for a maximum of $200$ epochs, with an early-stop mechanism ceasing training if there is no improvement in validation loss over $10$ consecutive epochs. 

The embedding dimensions $d_q$ and $d_c$ are set to 32. The learning rate, the hyperparameter $\lambda$, the number of layers in GNN of the concept map encoder $L$, and the embedding dimension $d_g$ are searched from [1e-3, 1e-4, 1e-5], [0.2, 0.5, 0.8], [1, 2, 3, 4], and [16, 32], respectively. For IRT-based prediction, $k$, for top-$k$ concepts, was tuned from the values [5, 7, 10, 20] to find the optimal setting that maximizes relevance accuracy. Additionally, the coefficients $\alpha$ and $\beta$ were explored within the range [0.01, 0.1, 1.0] to balance the contributions of the top-$k$ loss and contrastive learning loss to the overall model performance. 

\subsection{Prediction Performance (RQ1)}

\begin{table}[!t]
    \centering
    \caption{Accuracy (ACC) performance of all KT models on five datasets, where $*$ indicates a $\text{p-value} < 0.01$ in the t-test. The best-performing models are highlighted in \textbf{bold}, and the second-best models are {\ul underlined}.}
    \label{tab:main-acc}
    \renewcommand{\arraystretch}{1.3}
    \scalebox{0.75}{
        \begin{tabular}{l|>{\color{black}\arraybackslash}c >{\color{black}\arraybackslash}c >{\color{black}\arraybackslash}c >{\color{black}\arraybackslash}c >{\color{black}\arraybackslash}c}
            \hline
            \multicolumn{1}{c|}{\multirow{2}{*}{Model}} & \multicolumn{5}{c}{Dataset} \\ \cline{2-6} 
            \multicolumn{1}{c|}{} & Poly & DBE-KT22 & EdNet & NIPS34 & ENEM \\
            \hline
            GKT & 84.25±0.04 & 78.60±0.08 & 67.03±0.08 & 62.22±0.14 & 69.57±0.31 \\
            DKT & 84.79±0.07 & 79.05±0.02 & 67.87±0.03 & 65.07±0.04 & 72.16±0.01 \\
            SAKT & 84.75±0.10 & 79.14±0.12 & 68.61±0.10 & 69.36±0.06 & 73.08±0.02 \\
            AKT & {\ul 85.07±0.10} & 79.34±0.12 & {\ul 71.10±0.21} & {\ul 70.51±0.01} & 73.11±0.02 \\
            CL4KT & 84.67±0.07 & 79.20±0.08 & 67.87±0.01 & 69.35±0.04 & 73.16±0.02 \\
            DTransformer & 84.55±0.42 & 78.28±0.19 & 66.44±0.15 & 68.42±0.19 & 72.32±0.11 \\
            DP-DKT & 85.06±0.09 & {\ul 79.37±0.07} & 70.34±0.09 & 70.23±0.03 & {\ul 73.24±0.03} \\
            \hline
            \textbf{CRKT (ours)} & \textbf{85.75±0.12*} & \textbf{79.87±0.11*} & \textbf{71.69±0.12*} & \textbf{70.63±0.04*} & \textbf{73.54±0.02*} \\
            \hline
        \end{tabular}
    }
\end{table}

% Main Table (AUC)
\begin{table}[!t]
    \centering
    \caption{Area under the curve (AUC) performance of all KT models on five datasets, where $*$ indicates a $\text{p-value} < 0.01$ in the t-test. The best-performing models are highlighted in \textbf{bold}, and the second-best models are {\ul underlined}.}
    \label{tab:main-auc}
    \renewcommand{\arraystretch}{1.3}
    \scalebox{0.75}{
        \begin{tabular}{l|>{\color{black}\arraybackslash}c >{\color{black}\arraybackslash}c >{\color{black}\arraybackslash}c >{\color{black}\arraybackslash}c >{\color{black}\arraybackslash}c}
            \hline
            \multicolumn{1}{c|}{\multirow{2}{*}{Model}} & \multicolumn{5}{c}{Dataset} \\ \cline{2-6} 
            \multicolumn{1}{c|}{} & Poly & DBE-KT22 & EdNet & NIPS34 & ENEM \\
            \hline
            GKT & 70.51±0.87 & 78.41±0.25 & 63.32±0.18 & 66.95±0.16 & 69.48±0.18 \\
            DKT & 79.35±0.20 & 79.43±0.11 & 65.05±0.08 & 70.57±0.02 & 72.21±0.01 \\
            SAKT & 80.24±0.26 & 79.88±0.08 & 66.46±0.23 & 75.97±0.05 & 73.89±0.01 \\
            AKT & 81.00±0.12 & 80.02±0.07 & {\ul 72.18±0.26} & {\ul 77.19±0.02} & 74.17±0.03 \\
            CL4KT & 79.09±0.23 & 79.42±0.07 & 65.20±0.08 & 76.13±0.07 & 74.19±0.02 \\
            Dtransformer & 81.01±0.15 & 79.74±0.07 & 69.02±0.16 & 76.73±0.06 & 74.04±0.02 \\
            DP-DKT & {\ul 81.12±0.41} & {\ul 80.07±0.02} & 71.31±0.13 & 77.17±0.02 & {\ul 74.29±0.02} \\
            \hline
            \textbf{\mname{} (ours)} & \textbf{82.72±0.09*} & \textbf{81.04±0.06*} & \textbf{73.22±0.08*} & \textbf{77.27±0.03*} & \textbf{74.70±0.02*} \\
            \hline
        \end{tabular}
    }
\end{table}

We employ accuracy (ACC) and the area under the curve (AUC) as evaluation metrics for all datasets. Additionally, we conduct a t-test against the best baseline, where $*$ indicates $\text{p-value} <0.01$ in the t-test. According to Table~\ref{tab:main-acc} and Table~\ref{tab:main-auc}, our model \mname{} outperforms all baselines across the five datasets in both ACC and AUC, showing its reliability in students' correct or incorrect response predictions.

For the Poly, EdNet, and ENEM datasets, we observe that the performance of our model and DP-DKT, which utilizes option responses, is notably higher than that of other baselines. This observation suggests that content designers thoroughly consider options when designing questions. This is particularly credible given that the Poly dataset comprises English grammar questions, where the intention behind each incorrect option is likely more apparent than in other domains. Moreover, the EdNet and ENEM datasets consist of certified examinations managed by the Educational Testing Service and the Brazilian Ministry of Education, respectively, indicating a meticulous design of each option. The greater performance difference between DP-DKT and our model in the Poly dataset, compared to ENEM, is likely due to the presence of meticulously designed concepts and concept maps by the content designers.

Moreover, for DBE-KT22, a dataset designed and collected specifically for knowledge tracing, our model shows better results than other models. This implies that our model accurately encodes concept maps. The lowest performance gap is observed in the NIPS34 dataset, which is composed of crowd-sourced questions with a wide age range of participating students. This difference leads to inconsistent quality in questions and learning patterns, indicating that considering students' option responses in this context could act as noise.

Following the analysis, Table~\ref{tab:best-param} displays the optimal hyperparameters identified for each dataset. These settings were carefully adjusted to maximize the performance of \mname{} across various learning environments and evaluation metrics. 

\begin{table}[!t]
    \centering
    \caption{Best hyperparameters for each dataset}
    \label{tab:best-param}
    \renewcommand{\arraystretch}{1.2}
    \scalebox{0.9}{
    % \begin{tabular}{c|ccccc}
    \begin{tabular}{>{\color{black}\arraybackslash}c | >{\color{black}\arraybackslash}c >{\color{black}\arraybackslash}c >{\color{black}\arraybackslash}c >{\color{black}\arraybackslash}c >{\color{black}\arraybackslash}c}
        \hline
        Dataset & Poly & DBE-KT22 & EdNet & NIPS34 & ENEM \\ \hline
        learning rate & 1e-3 & 1e-3 & 1e-3 & 1e-3 & 1e-3 \\
        $\lambda$ & 0.5 & 0.5 & 0.2 & 0.2 & 0.2 \\
        $L$ & 3 & 3 & 3 & 2 & 2 \\
        $d_g$ & 32 & 32 & 32 & 32 & 16 \\
        $k$ & 7 & 10 & 10 & 10 & 5 \\
        $\alpha$ & 0.1 & 0.1 & 0.01 & 0.01 & 0.01 \\
        $\beta$ & 0.1 & 0.1 & 0.1 & 0.01 & 0.1 \\ \hline
    \end{tabular}
    }
\end{table}

% Ablation Study
\subsection{Ablation Study (RQ2)}
To explore the individual impact of each module in \mname{}, we design four variations and perform an ablation study. Each variant is as follows:

\begin{itemize}
    \item \textbf{noOpt} removes the option responses and instead uses binary responses like other KT models.
    \item \textbf{noUnc} removes the unchosen response, utilizing only the chosen option response.
    \item \textbf{noMap} removes the concept map, utilizing the concept-level knowledge state directly extracted from the knowledge retriever.
    \item \textbf{noTopK} allows the knowledge states of all concepts to contribute to the ability calculation.
\end{itemize}

% [Figure] Ablation Study
\begin{figure}[!ht]
    \centering
    \includegraphics[width=0.9\textwidth]{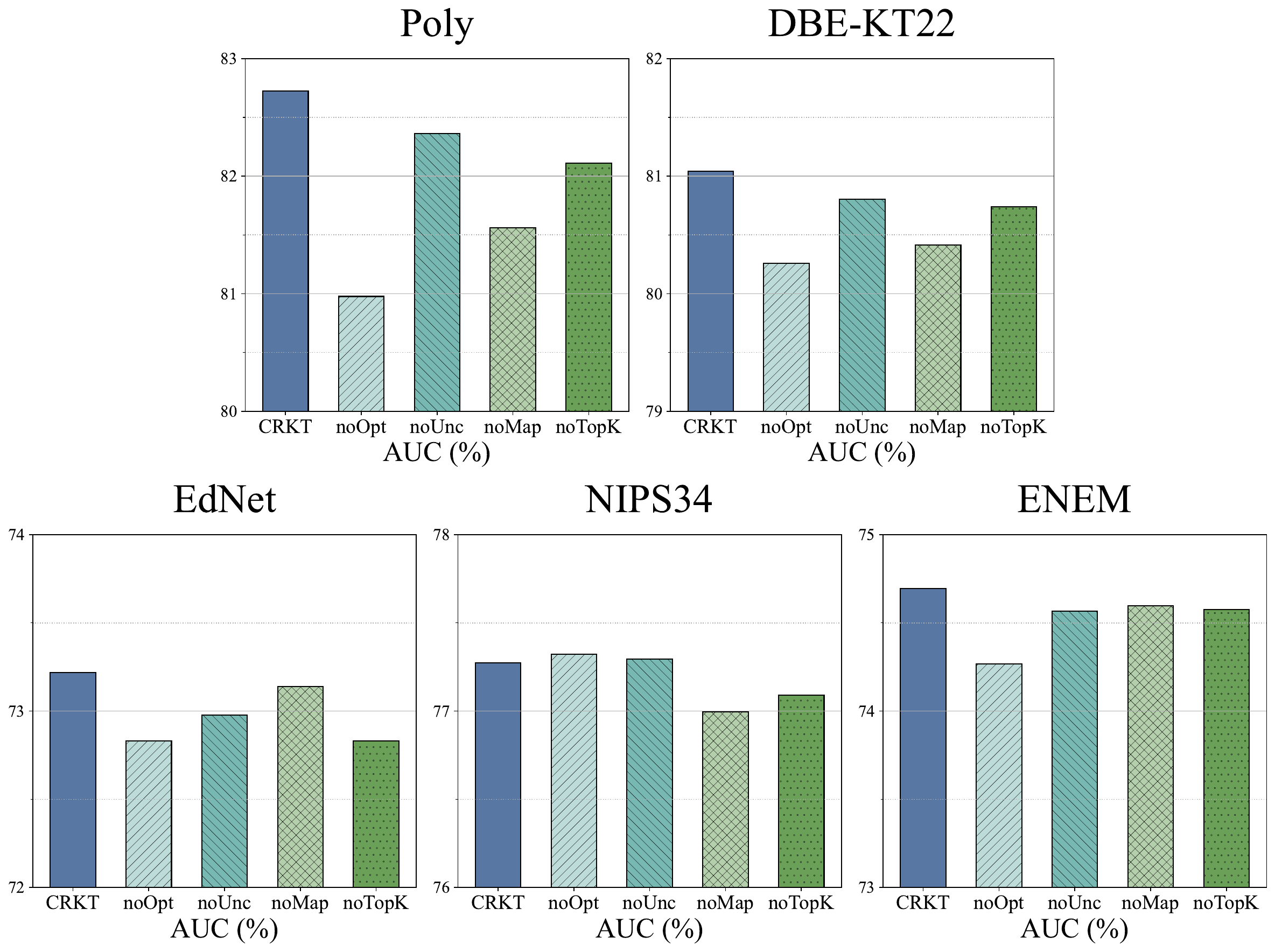}
    \caption{The experimental results in ablation study.}
    \label{fig:ablation}
\end{figure}

Fig.~\ref{fig:ablation} presents the average results of the 5-fold cross-validation across all datasets, revealing that all variations, except for noOpt and noUnc in the NIPS34 dataset, reduce \mname{}'s performance. For the Poly dataset, removing option responses (noOpt) leads to the most significant performance drop in both metrics, implying the importance of considering students' detailed option responses in enhancing model performance. This trend is consistent across the DBE-KT22, EdNet, and ENEM datasets, indicating that option responses critically contribute to the model's performance in accurately predicting students' knowledge states. In contrast, no performance decline exists for noOpt and noUnc in the NIPS34 dataset. This shows that the option responses might not contribute positively to estimating the knowledge state across a wide range of student ages, possibly due to the heterogeneity of the questions or the participants' responses. Removing the concept map (noMap) also has a significant performance drop in all datasets except for ENEM and EdNet. This indicates the importance of concepts and their relationships within the concept map defined by experts.

These findings from the ablation study clearly illustrate the essential roles that option responses and concept maps play in the \mname{}'s ability to predict students' knowledge states accurately. The variations in performance across different datasets also provide insights into how certain features may vary in their importance based on the dataset's characteristics, such as the age range of the students or the design of the questions and concept map.

% text 정보 활용 비교
\subsection{Usage of Textual Information (RQ3)}
We investigate the effectiveness of utilizing textual information within the proposed \mname{} model, which typically derives question and response representations through learning-based approaches. We aim to determine how effectively the dataset, which provides detailed information about questions and options, can leverage textual information from the question's text.

Given the inherent challenge of distinguishing between option representations directly from their text, we utilize the capabilities of a large language model (LLM) to enhance our representations. Specifically, we provided the question information to OpenAI's ChatGPT\footnote{\url{https://openai.com/index/chatgpt}}, which then generated explanations for the solution approach and reasons why a student might choose each option. These generated texts were then employed as representations for questions and each option's responses.

\begin{table}[!ht]
    \centering
    \color{black}
    \caption{Comparison of AUC performance based on different representations}
    \label{tab:text}
    \renewcommand{\arraystretch}{1.3}
    \scalebox{0.9}{
    \begin{tabular}{c|cc}
        \hline
        Dataset & Poly & NIPS34 \\
        \hline
        learning-based & 82.72±0.09 & 77.27±0.03 \\
        textual information & 82.96±0.06 & 77.25±0.02 \\
        \hline
    \end{tabular}
    }
\end{table}

Our experiments were conducted on the Poly and NIPS34 datasets, and the results are shown in Table~\ref{tab:text}. The results do not demonstrate significant performance improvements in either dataset. This suggests that the benefit of using textual information, given the cost of obtaining these representations, is marginal. 
% Consequently, we opt to continue using learning-based representations within \mname{}.

% Concept Map Analysis
\subsection{Performance via Correct Rate (RQ4)}

% [Figure] Performance via Correct Rate
\begin{figure}[!ht]
    \centering
    \includegraphics[width=0.9\textwidth]{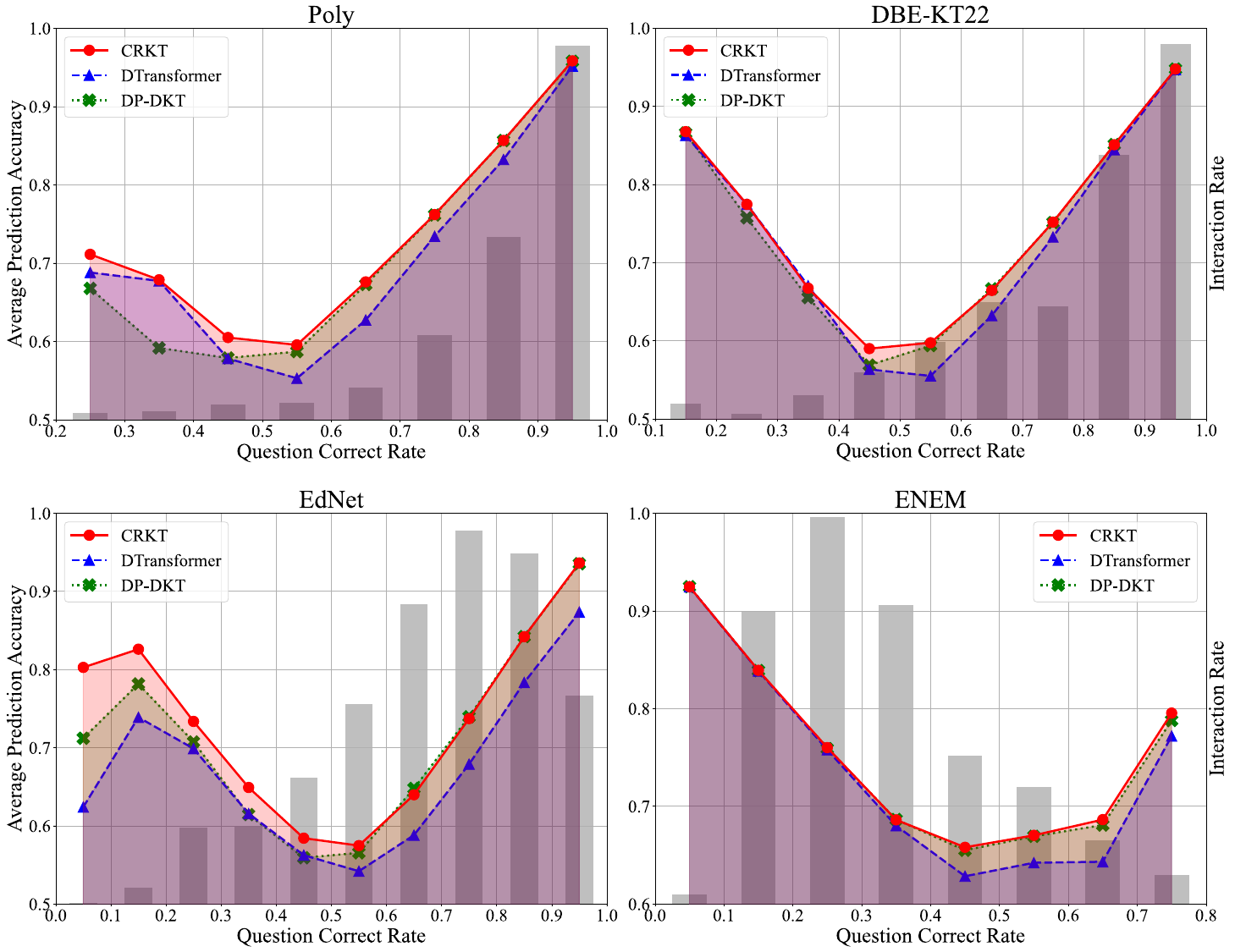}
    \caption{Comparison of accuracy between DTransformer, DP-DKT, and \mname{} across different average question correctness rates(x-axis) and interaction rate(bar).}
    \label{fig:correct_rate}
\end{figure}

In this section, we analyze the performance of different models across four datasets based on the average correct rate of questions. As depicted in Fig.~\ref{fig:correct_rate}, we observe that model accuracy tends to increase at the extremes of the average correct rate spectrum. This trend likely stems from the skewed distribution of score labels for questions at these extremes. Notably, the models struggle most with questions with an average correct rate between 40\% and 60\%, indicating these are the most challenging for prediction.

Among the baseline models, DP-DKT shows the highest prediction performance. However, in comparison to DTransformer, which employs contrastive learning, \mname{} demonstrates superior performance, particularly for questions with an average correct rate within the 40\% to 60\% range. This suggests that \mname{} effectively distinguishes the knowledge state of students tackling these moderately difficult questions, likely benefiting from well-designed response flip augmentation that enhances its contrastive learning component. Moreover, \mname{} maintains robust prediction performance even in data-sparse regions, further evidencing its capability to track the knowledge state across varying difficulties accurately.

\subsection{Case Study (RQ5)}

% [Figure] Case Study
\begin{figure}[!ht]
    \centering
    \includegraphics[width=\textwidth]{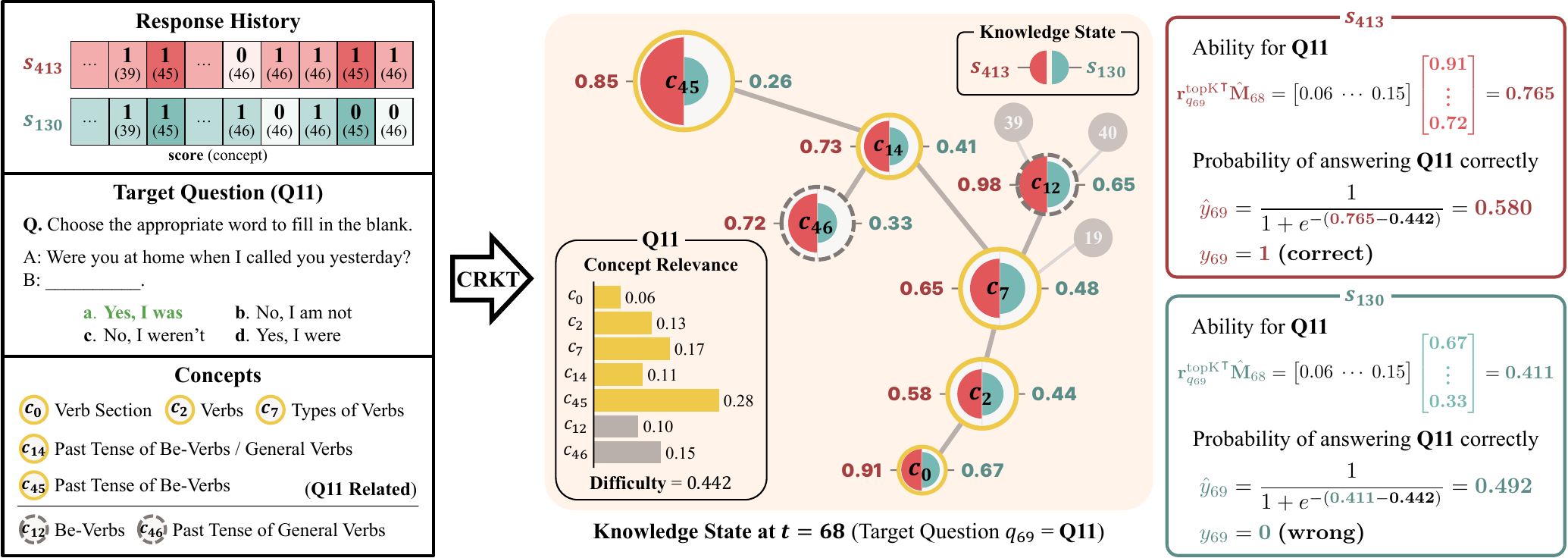}
    \caption{Visualization of knowledge states and prediction extracted by \mname{} for two students with identical question histories solving the same target question (Q11) from the Poly dataset.}
    \label{fig:case}
\end{figure}

We conduct a visualization analysis to validate the representational capabilities and interpretability of \mname{}'s knowledge states. Fig.~\ref{fig:case} above illustrates the knowledge states and prediction process extracted by \mname{} for two students with identical question histories tackling the same target question (Q11) in the Poly dataset.

The concepts directly related to the target question (Q11), highlighted with a yellow border, are consistently identified among the top-7 relevant concepts extracted by \mname{}. Neighboring nodes in the concept map, $c_{12}$ and $c_{46}$, are also identified. Considering that the question pertains to the past tense of be-verbs, the concepts `Be-Verbs' and `Past Tense of General Verbs' also play a significant role in predicting the correctness of the response. This underlines the relevance of these concepts in assessing the student's grasp of the subject matter.

Student $s_{413}$ demonstrates a better mastery of concepts related to the target question compared to student $s_{130}$, as indicated by their response history to related questions. This difference in concept mastery contributes to the variance in their ability values for the target question Q11, as calculated by the IRT-based prediction mechanism, leading to different predicted outcomes for each student.

This case study exemplifies how \mname{} can provide meaningful interpretations in real-world scenarios, showcasing its potential to tailor educational content and assessments to individual learning trajectories.

    %!TEX root = 0_main.tex

\section{\rv{Discussion and Conclusion}}

\subsection{\rv{Discussion}}
\mname{} meets the research objective of enabling the practical application of KT models in educational services. However, there are still limitations that need to be addressed.

A limitation of \mname{} is its reliance on predefined concept maps to accurately encode the correct relationships among concepts. In this paper, we employ statistical methods to infer these relationships in datasets lacking such concept maps. However, the effect of these inferred relationships is not as large as expected, suggesting that the effectiveness of the currently employed statistical approach could be improved. Enhancing these methods could improve our model's accuracy and efficiency by better capturing the interdependencies among concepts.

For future research, we aim to refine \mname{} to ensure it can provide stable, reliable, and interpretable feedback within actual educational services. This includes improving the model's ability to integrate and utilize concept relationships without the need for predefined maps and advancing the statistical methods used to determine these relationships. 
For example, the work in \cite{li2024automate} preliminarily proposed a method to tag knowledge concepts for questions with Large Language Models (LLMs). Authors in \cite{dagdelen2024structured} also suggested the potential usefulness of structuring textual data by extracting hierarchical relationships among concepts.
By developing more sophisticated approaches to identify concept interactions dynamically, we hope to create a more versatile and universally applicable model that can function effectively across diverse educational datasets without the need for manual input of concept maps. These advancements will support \mname{}'s broader application in real-world educational settings, enhancing personalized learning experiences through more accurate and meaningful feedback.

\subsection{\rv{Conclusion}}
In this paper, we introduce the \mfull{} (\mname{}) model, a new approach to knowledge tracing that enhances prediction performance and interpretability by utilizing student responses and concept maps. With the novel use of unchosen MCQ options and concept-level knowledge through the concept map, \mname{} provides a more accurate update to knowledge states, offering precise and actionable feedback for personalized learning. Our extensive experiments across multiple datasets demonstrate \mname{}'s superiority in predicting student performance and its practicality in real-world educational settings. Future research will aim to refine the \mname{} model to offer reliable, interpretable feedback for seamless integration into actual educational services, focusing on enhancing its applicability and effectiveness in everyday educational scenarios.

    \section*{Acknowledgements}
    % This work was supported by Institute of Information \& communications Technology Planning \& Evaluation(IITP) grant funded by the Korea government(MSIT) (No.2019-0-00421, Artificial Intelligence Graduate School Program(Sungkyunkwan University))
    This work was also supported by the Institute of Information \& Communications Technology Planning \& Evaluation (IITP) grant funded by the Korea government (MSIT): (No. 2019-0-00421, Artificial Intelligence Graduate School Program (Sungkyunkwan University)), (No. RS-2023-00225441, Knowledge information structure technology for the multiple variations of digital assets), and (No. RS-2024-00438686, Development of software reliability improvement technology through identification of abnormal open sources and automatic application of DevSecOps). This research was also supported by Culture, Sports, and Tourism R\&D Program through the Korea Creative Content Agency grant funded by the Ministry of Culture, Sports and Tourism in 2024 (Project Name: Development of technology for dataset copyright of multimodal generative AI model, Project Number: RS-2024-00333068, Contribution Rate: 20\%). This research was also supported financially by the grant from the National Research Foundation of Korea (NRF-2021M3H4A1A02056037). 

    \bibliographystyle{abbrvnat}
    \bibliography{7_references}

\end{document}